\documentclass[10pt,twocolumn,letterpaper]{article}
\pdfoutput=1
\usepackage{cvpr}
\usepackage{times}
\usepackage{epsfig}
\usepackage{graphicx}
\usepackage{amsmath}
\usepackage{amssymb}
\usepackage{multirow}
\usepackage{algorithm}
\usepackage[noend]{algpseudocode}
\usepackage{balance}
\usepackage{tabularx}
\usepackage{array}
\makeatletter
\def\BState{\State\hskip-\ALG@thistlm}
\makeatother

\cvprfinalcopy 

\def\cvprPaperID{1408} 

\ifcvprfinal\fi
\begin{document}

\title{Learning Markov Clustering Networks for Scene Text Detection}

\author{Zichuan Liu$^1$, Guosheng Lin$^1$, Sheng Yang$^1$, Jiashi Feng$^2$, Weisi Lin$^1$ and Wang Ling Goh$^1$\\
	$^{1}$Nanyang Technological University, Singapore\\
	$^{2}$National University of Singapore, Singapore\\
	{\tt\small \{zliu016, syang014\}@e.ntu.edu.sg}, {\tt\small \{gslin, wslin, ewlgoh\}@ntu.edu.sg} \\
	{\tt\small elefjia@nus.edu.sg}
}

\maketitle
\thispagestyle{empty}

\begin{abstract}

A novel framework named Markov Clustering Network (MCN) is proposed for fast and robust scene text detection. MCN predicts instance-level bounding boxes by firstly converting an image into a Stochastic Flow Graph (SFG) and then performing Markov Clustering on this graph. Our method can detect text objects with arbitrary size and orientation without prior knowledge of object size. The stochastic flow graph encode objects' local correlation and semantic information. An object is modeled as strongly connected nodes, which allows flexible bottom-up detection for scale-varying and rotated objects. MCN generates bounding boxes without using Non-Maximum Suppression, and it can be fully parallelized on GPUs. The evaluation on public benchmarks shows that our method outperforms the existing methods by a large margin in detecting multioriented text objects. MCN achieves new state-of-art performance on challenging MSRA-TD500 dataset with precision of 0.88, recall of 0.79 and F-score of 0.83. Also, MCN achieves realtime inference with frame rate of 34 FPS, which is $1.5\times$ speedup when compared with the fastest scene text detection algorithm. 

\end{abstract}
\section{Introduction}

Detecting structural objects in an image is a ubiquitous problem in real-word. Powered by the recent advances in Convolutional Neural Networks (CNNs), the object detection system has achieved human-level accuracy with real-time processing capability \cite{girshick2015fast,ren2015faster,redmon2016you,liu2016ssd,ding2018cvpr}. Despite the progresses made in general object detection, we still confront problems in detecting objects in a specific application area. \par

\begin{figure}[]
	\centering
	\includegraphics[width=0.9\linewidth]{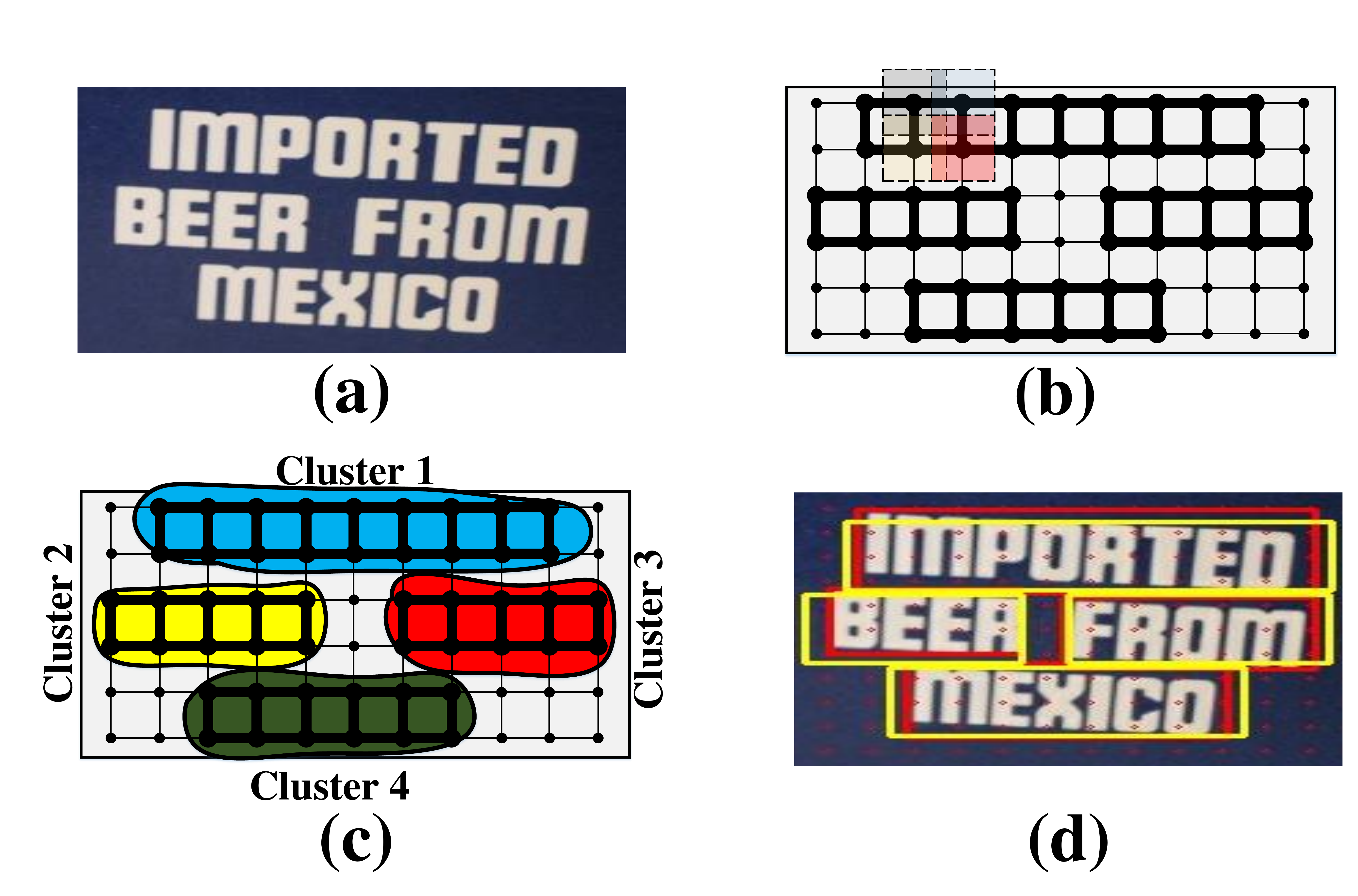}
	\caption{(a) Input image; (b) Predicted stochastic flow graph by MCN: Nodes correspond to the equidistant overlapping regions in the image and the connections between nodes are refered as flows. The flow intensity is visualized as the width of an edge. A strong flow results in a wide edge while weak flow leads to a narrow edge; (c) Extracted clusters from stochastic flow graph presented in (b) by Markov Clustering; (d) Bounding boxes generated from the clustered nodes in (c).}
	\label{fig:system_flow}
\end{figure}

In scene text detection, existing CNN-based methods may fail when producing bounding boxes with extremely large aspect ratio or unsupported orientation \cite{tian2015text,shi2017detecting}. These methods \cite{ren2015faster,redmon2016you,liu2016ssd} follow the top-down prediction paradigms, where object boxes are produced by appreciating the global information of an object while neglecting the local information. Therefore, the top-down method usually requires prior knowledge of the text box geometry to design reference boxes, which is task-specific and heuristic. As a result, to maintain the detection performance for various text sizes and orientations, one will inevitably increase the number of reference boxes, and thus lower the inference speed due to the increased output dimension \cite{liao2017textboxes}. On the other hand, due to the absence of the local semantic information, the existing methods have to rely on Non-Maximum Suppression (NMS) \cite{neubeck2006efficient} to remove redundant bounding boxes, which is unparallelizable on GPUs. 

To address these issues, we propose an unified framework called Markov Clustering Network (MCN) for detecting scale-varying and arbitrarily oriented texts. It is an end-to-end trainable model describing both the local correlation and semantic information of an object with Stochastic Flow Graph (SFG). As shown in Figure \ref{fig:system_flow}, equidistant and overlapping regions are considered as nodes of SFG with edges weighted by flow values. Nodes belonging to the same object are strongly connected by the flows and will be grouped together by applying fully paralleled Markov Clustering (MC) on the SFG. Bounding boxes are produced based on the generated clusters with post-processing. \par 

In contrast with the top-down methods \cite{ren2015faster,redmon2016you,liu2016ssd}, our method predicts bounding boxes in a bottom-up manner. Essentially, the MCN predicts instance-level objectness by merging the dense object predictions according to the local correlation measurements. This framework can naturally detect texts with arbitrary size and orientation. Our method does not use NMS to produce bounding boxes and can be fully parallelized on GPUs.

We evaluate our method on public benchmarks and prove its robustness to large variation of scale, aspect ratio and orientation. Our method achieves the state-of-art performance with much faster inference. The contribution of this work is summarized as follows:
\begin{itemize}
	\item A bottom-up method for scene text detection is proposed which assembles local predictions into object bounding boxes by performing Markov Clustering on Stochastic Flow Graph;
	\item Markov Clustering is regarded as a set of special differentiable neural network layers and an end-to-end training method is developed for learning graph clusters from image data;
	\item The proposed inference process is fully paralleled on GPUs and achieves realtime processing capability with frame rate of 34 FPS, which means to $1.5\times$ speedup when compared with fastest scene text detection algorithm.
	\item Our method outperforms existing scene text detection methods in detecting arbitrarily oriented text objects, and achieves new state-of-art performance on challenging MSRA-TD500 dataset with precision of 0.88, recall of 0.79 and F-score of 0.83. 
\end{itemize}
\section{Related Works} 
Over the past few years, much research effort have been devoted to text detection at character level \cite{neumann2016real,wang2012end,huang2013text,huang2014robust} and word level \cite{yao2012detecting,wang2010word,zhang2016multi,zhang2015symmetry,bissacco2013photoocr,jaderberg2016reading,gupta2016synthetic}. Character-based methods detect individual characters and group them into words. These methods find characters by classifying candidate regions extracted by region extraction algorithms or by classifying sliding windows. Such methods often involve a post-processing step of grouping characters into words. Word-based methods directly detect word bounding boxes. They often have a similar pipeline to the recent CNN-based general object detection networks. \par

Recently, the segment-based method has opened up a new direction to solve this problem \cite{tian2016detecting,shi2017detecting}. Instead of detecting the whole object, these methods target at detecting segments of an object and combining these segments to a bounding box. Work \cite{tian2016detecting} combines spatial recurrent components with YOLO architecture to detect segments and connects the segments heuristically according to their horizontal distance. Inheriting from the SSD \cite{liu2016ssd} method,  \cite{shi2017detecting} predicts both object segments and links in between on multi-resolution feature maps. Instance-level bounding boxes are generated by merging oriented bounding boxes according to the link scores between them. However, this method still requires predefined default box for bounding box regression, and excessive connections between segments significantly complicates the training and slows down the inference. \par

Different from the existing methods, our method treats detection as a graph clustering problem. Instance-level object regions are represented by strongly connected nodes in a graph which can be extracted by Markov Clustering. Therefore, our method can generate bounding boxes with arbitrary box geometry.\par

\section{Method}
\subsection{Overview}

Markov Clustering Network (MCN) is an object detection method based on graph clustering. An $H\times W$ image is translated by MCN into a spatial feature map which will be further constructed into a latticed graph $G(V,E)$ called Stochastic Flow Graph (SFG). The nodes $V$ in $G$ correspond to the feature vectors extracted from the overlapping regions of the image. The edges $E$ are weighted by the flow values $f_0$, $f_1$, $f_2$ and $f_3$ predicted by MCN. They are 2D maps with size of $\frac{H}{U}\times\frac{W}{U}$ denoting the connection intensity or interaction to current node or its three neighbors. In our prediction framework, the presence of an object is jointly represented by nodes with strong connections to each other, and the background region is represented by isolated nodes. Therefore, detecting an object is equivalent to predicting the flow values and then grouping the nodes according to their connection intensities. Given the flow values predicted by MCN, we extract the objectness by performing Markov Clustering (MC) \cite{van2001graph} on $G$. The strong connected nodes are grouped into clusters representing objects. By mapping the nodes of a cluster back to the input image, the corresponding bounding boxes can be produced by simple post-processing.

\subsection{Object Representation by Stochastic Flow} \label{sect:obj_rep_by_stoc_flow}

The existing object detection methods can be categorized as top-down methods, where the detection relies on coarse global observation of an image \cite{liu2016ssd,redmon2016you,ren2015faster}. Due to the absence of the local information, these methods usually predict offset of the object size and orientation relative to predefined references (reference boxes) \cite{liao2017textboxes,shi2017detecting}. Designing these references is task-specific which can hardly cover all cases and will degrade the detection robustness. This problem is getting worse in detecting objects with arbitrary aspect ratio and various orientation. If the object geometry is not well-supported by the references, large amounts of failure will occur when detecting these objects. \par

\begin{figure}[]
	\centering
	\includegraphics[width=\linewidth]{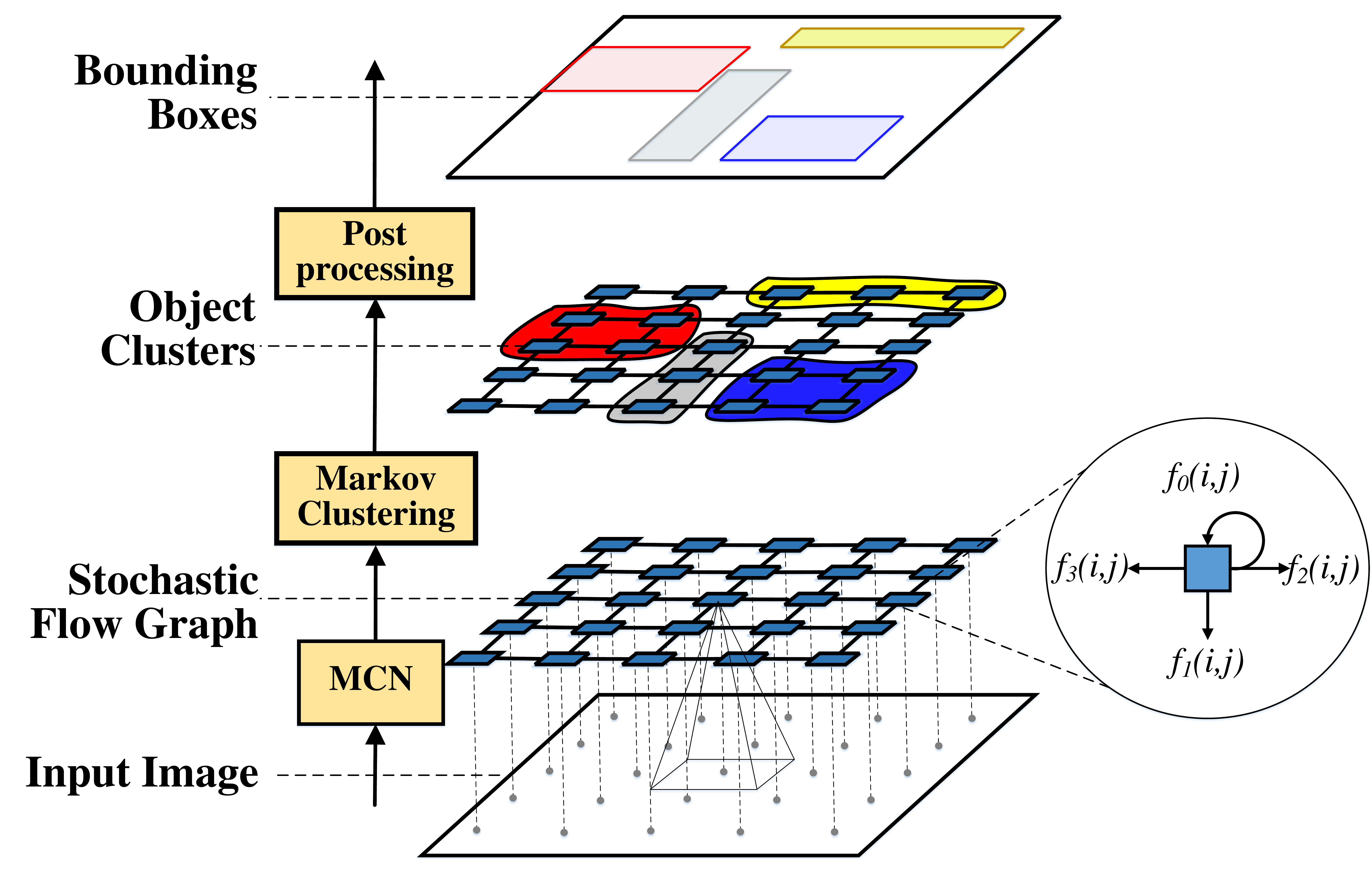}
	\caption{Markov Clustering Network.}
	\label{fig:method}
\end{figure}

Our method considers the object detection in a bottom-up manner to solve the problems mentioned above. In our method, as shown in Figure \ref{fig:method}, an input image is converted via MCN to a Stochastic Flow Graph (SFG) $G$ with nodes $V$ and directed edges $E$ weighted by stochastic flows $f_0$, $f_1$, $f_2$ and $f_3$. For a given node $V(i,j)$, the corresponding flows $f_0(i,j)$, $f_1(i,j)$, $f_2(i,j)$ and $f_3(i,j)$ are positive and sum up to 1. An object is abstracted as nodes connected by the outgoing flows $f_1$, $f_2$ and $f_3$, while the background region is represented by nodes isolated by the self-loop flows $f_0$. Since the nodes have corresponding spatial relation in the original image, the presence as well as the geometry (size and orientation) of an object can be represented by nodes and their flows, which is insensitive to variation of size and orientation. \par 

From the point of probability, the SFG is actually modeling the Markov random walk process, where each node denotes a state in a Markov chain and the corresponding directed weighted edges represent the transition probabilities of this state. For a random walk process starting at a given node $V(i,j)$, there exists a stationary distribution (or flow distribution) $P(V|V(i,j))$ describing possible destination nodes of this process. Specifically, the node with maximum value in $P(V|V(i,j)$ is denoted as the attractor of $V(i,j)$. Therefore, the strongly connected nodes can be regarded as nodes with the same attractor \cite{van2001graph}. This interpretation provides us a probabilistic description of flows and clusters. Moreover, it allows us to uniquely represent an instance-level object region with an attractor, which is the fundamental of our detection method. \par
 
\subsection{Detecting Object by Markov Clustering}

Based on the probabilistic interpretation and property of SFG, we apply Markov Clustering (MC) to extract the instance-level object regions. Markov Clustering is an algorithm to identify the strongly connected nodes and group them into clusters. In Markov Clustering, a flow matrix $M_0$ is constructed from $G$ with entry $M_0(m,n)$ representing the flow value from node $V(i_n,j_n)$ to node $V(i_m,j_m)$ \footnote{We use both 1D and 2D notation, alternatively, to index a node. The transformation between 1D notation $m$ and 2D notation $(i_m,j_m)$ can be represented by $m = i_m + \frac{H}{U}\cdot j_m$.}. The $n$-th column $M_0(:,n) \in \mathbb{R}^{\frac{H}{U}\cdot\frac{W}{U}}$ of $M_0$ represents the transition probability of a Markov random walk starting at node $V_n$, which is denoted as $P_0(V|V(i_n,j_n))$. Markov Clustering is actually computing the stationary distribution $P(V|V(i,j))$ for each node. It consists of a set of iterations including matrix-matrix multiplication and non-linear transformation, which are illustrated as follows:

\textbf{Expand}: Input $M_{t-1}$, output $M_t$.
\begin{equation}
M_t = M_{t-1}*M_0
\end{equation}

\textbf{Inflate}: Input $M_t$, output $M_t$.
\begin{equation}
M_t(m,n) = \frac{M_t(m,n)}{\sum_l{M_t(l,n)}}
\end{equation}

\textbf{Prune}: Input $M_t$, output $M_t$.
\begin{equation}
M_t(m,n) = 0,\ if \ M_t(m,n) < \emph{threshold}
\end{equation}
where $M_t$ is the intermediate result at $t$-th iteration and $N$ is the number of iterations for convergence. The expansion step spreads the flows out of a node to its potential new node. It enhances the flows to the nodes which are reachable by multiple paths. The inflation step and pruning step are meant to regularize the iteration to ensure convergence by introducing a non-linearity into the process, while also have the effect of strengthening intra-cluster flows and weakening the inter-cluster flows \cite{van2001graph}. The pseudo-code for Markov Clustering in presented in Algorithm \ref{alg:mc}.


\begin{algorithm}
	\caption{Markov Clustering}
	\label{alg:mc}
	\begin{algorithmic}[1]
		\BState {\textbf{Initialize}:}
			$t=0$\;
		\While{$t < N$} 
			\State{$M_t = Expand(M_{t-1})$\;} 
			\State{$M_t = Inflate(M_t)$}\;
			\State{$M_t = Prune(M_t)$}\;
		\EndWhile
		\BState {\textbf{Output}: $M_N$ as a clustering.}
	\end{algorithmic}
\end{algorithm}
At the start of the process, the outgoing flow distribution of a node is smooth and uniform, and becomes more and more peaked as the iterations are executed. The columns of $M_t$ corresponding to the same cluster will converge to the same one-hot vector. It is reflected on $G$ that nodes within a tightly-linked group will flow to the same attractor at the end, which helps to identify any potential cluster. In addition, Markov Clustering does not require predefined number of clusters, and due to the parallelizability of three operations, Markov Clustering can be fully parallelized on GPUs. \par

\subsection{Learning Clustering with Flow Labels}

In this section, we illustrate the learning algorithm for MCN to correctly predict the stochastic flow for clustering nodes. \par

\textbf{Locating Attractors for Clusters}
As illustrated previously, the converged flow matrix $M_N$ describes the flow distribution of possible attractors for each node. Therefore, labeling clusters is equivalent to labeling attractor for each node. 
\begin{figure}[]
	\centering
	\includegraphics[width=\linewidth]{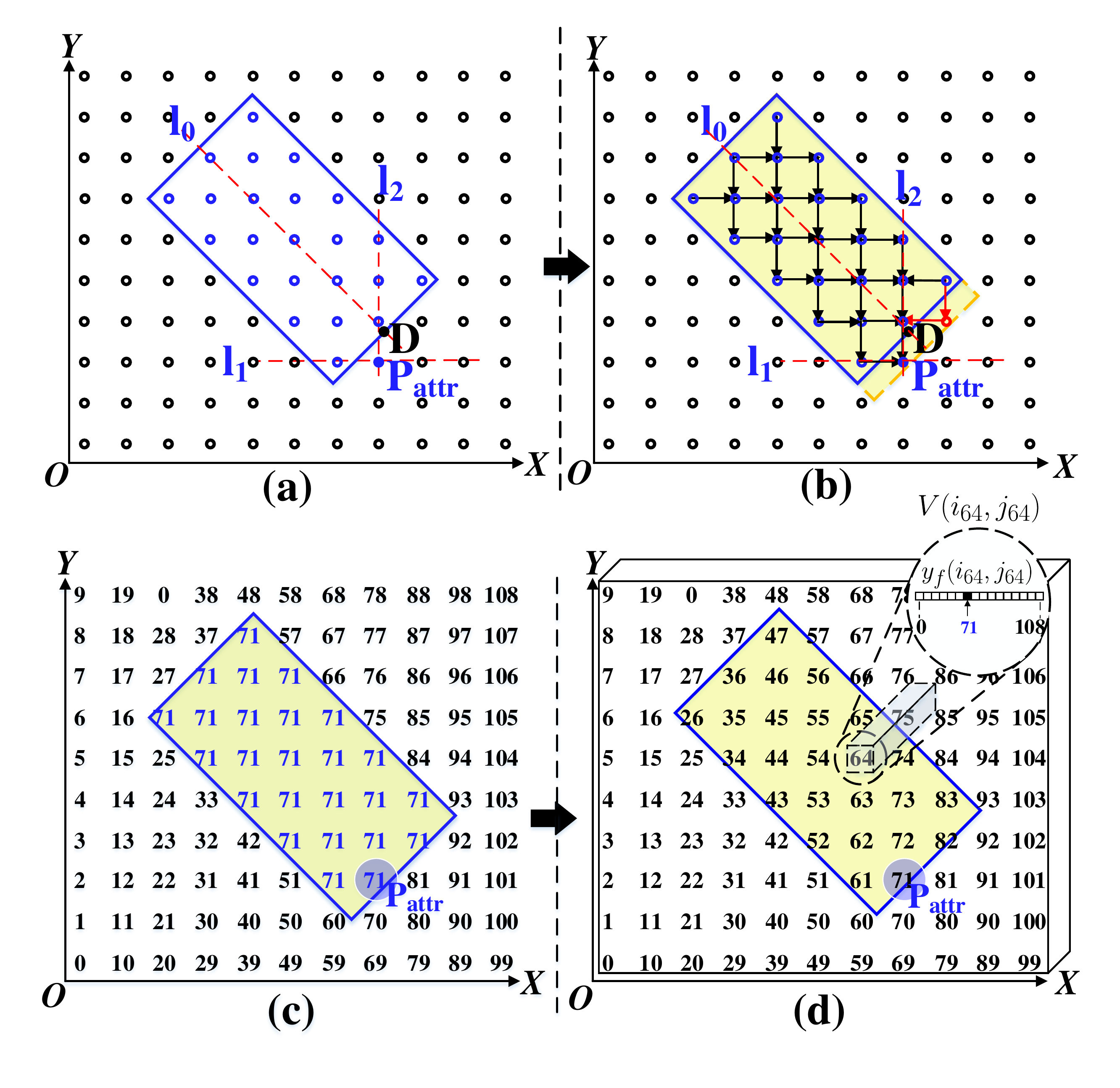}
	\caption{(a) Computing the attractor $P_{attr}$ given a ground-true bounding box; (b) Adjusting a ground-true bounding box to include the attractor; (c) Assigning an attractor index for each node; (d) Generating 3D cluster label for each node based on (c).}
	\label{fig:labelling}
\end{figure}
Defining the nodes within the same bounding box as a cluster, we compute the attractor for this cluster based on the geometry of the ground-true bounding box. As shown in Figure \ref{fig:labelling} (a), given a ground-truth bounding box, we firstly compute the coordinates of $D$, which is the intersection between the major axis and the lower short-side of the bounding box. Second, we draw a horizontal line $l_1$ that traverses the node with lowest $Y$-coordinates in the bounding box region, and a vertical line that $l_2$ traverses the nearest node from $D$. Finally, the intersection node between $l_1$ and $l_2$ is determined as the attractor. To ensure attractor being in a bounding box, we adjust the bounding box size, which may introduce new nodes into it. \par

\textbf{From Attractors to Cluster Labels} The Markov Clustering outputs the stationary distribution $M_N(:,m) \in \mathbb{R}^{\frac{H}{U}\cdot\frac{W}{U}}$ of potential attractors for each node. Thus, the ground-true label for each node is defined as the target distribution $y_f(i_m,j_m) \in \mathbb{R}^{\frac{H}{U}\cdot\frac{W}{U}}$. As shown in Figure \ref{fig:labelling} (c), we firstly make 2D mask to record the 1D attractor index $m\in [0,\cdots,\frac{H}{U}\cdot\frac{W}{U}-1]$ for each node. For the nodes within an object region, they share the same attractor index, while for an node corresponding to the background, it becomes the attractor of itself. Based on the attractor mask, we generate a 3D cluster (flow) label $y_f\in \mathbb{R}^{\frac{H}{U}\times\frac{W}{U}\times(\frac{H}{U}\cdot\frac{W}{U})}$ describing the target stationary distribution for all nodes, which is shown in Figure \ref{fig:labelling} (d). For specific node $V(i_m,j_m)$ with attractor $V(i_k, j_k)$, the target distribution $y_f(i_m,j_m)$ is a one-hot vector with $k$-th entry labeled as 1.\par

\textbf{Loss Function} Given the converged flow distribution $M_N(:,m)$ for a node $V(i_m, j_m)$ and the target distribution $y_f(i_m, j_m)$, the loss function is represented by a cross-entropy loss between these two distribution:
\begin{align}
\begin{split} L_f(i_m, j_m) &= -y_f(i_m, j_m)\cdot \ln(M_N(:,m)), \end{split} 
\end{align}
and the flows of all nodes are globally optimized by minimizing the mean cross-entropy error represented by:
\begin{align}
\begin{split} C_f &= \frac{1}{H/U\cdot W/U}\sum_{i_m}\sum_{j_m} L_f(i_m, j_m) \end{split} 
\end{align}

\begin{figure}[]
	\centering
	\includegraphics[width=\linewidth]{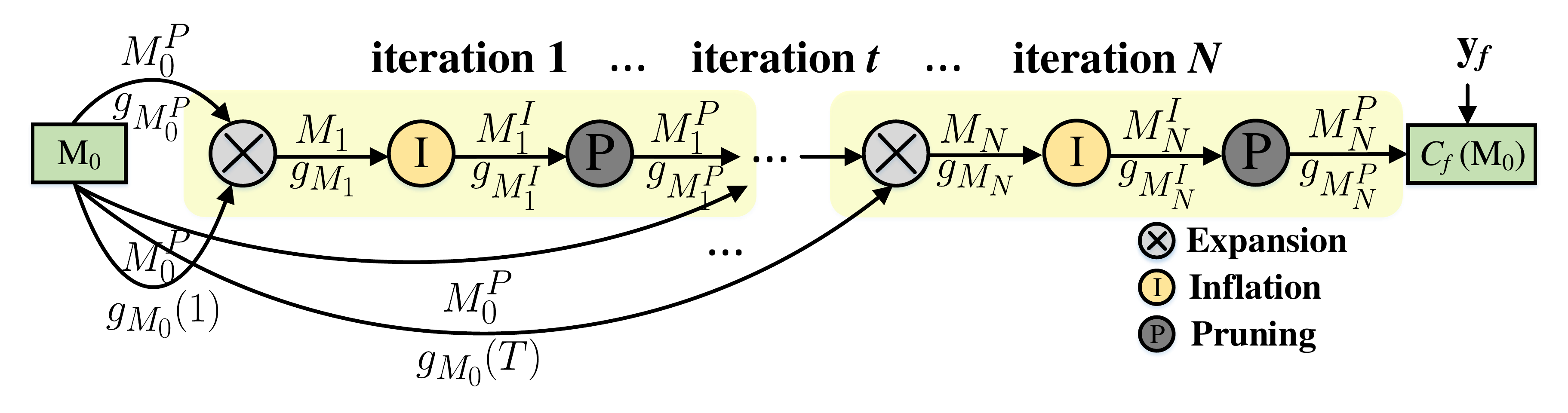}
	\caption{Computing graph of Markov Clustering. It takes initial flow matrix $M_0$ as input and outputs a converged flow $M_N^P$ to compute $C_f(M_0)$.}
	\label{fig:flow_grad}
\end{figure}
\begin{figure*}[]
	\centering
	\includegraphics[width=0.9\linewidth]{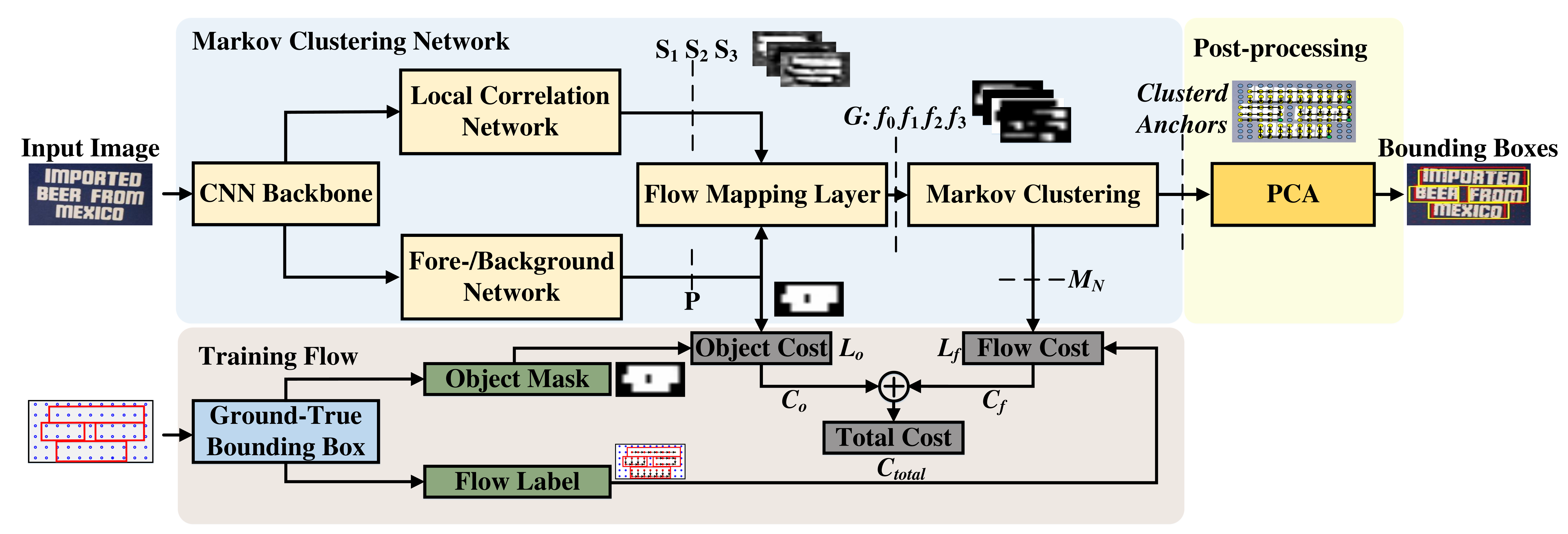}
	\caption{Implementation detail of MCN for scene text detection.}
	\label{fig:sys_flow}
\end{figure*}

\textbf{Gradients of Markov Clustering} 
An end-to-end supervised training requires the differentiability of all the operations in a model and the feasibility of labeling the data. In this section, we focus on the differentiability of Markov Clustering. The operations included in Markov Clustering can be treated as special neural network layers, which are differentiable. We visualize the operations with a computing graph which computes the stationary distribution $M_N$ and corresponding cross-entropy loss $C_f(M_0)$ given flow matrix $M_0$ and target distribution $y_f$. In Figure \ref{fig:flow_grad}, each node represents one operation in Markov Clustering, and the directed edges show the data flow throughout the whole clustering process for $N$ iterations. The output data of an operation is marked above the edge and corresponding gradient $g_{(\cdot)}$ is marked below. From the computing graph, gradient of cost function of stochastic flow $C_f$ respecting $M_0$ is derived by using the chain rule illustrated below:
\begin{align}
\label{eq:flow_grad}
\begin{split} &g_{M_N^P} = \partial{C_f}/\partial{M_N^P}, \end{split} \\
\begin{split} &g_{M_t^I} = g_{M_t^P}\cdot f'_P(M_t^I),  \end{split} \\
\begin{split} &g_{M_t} = g_{M_t^I}\cdot f'_I(M_t),  \end{split} \\
\begin{split} &g_{M_{t-1}^P} = M_0^T\cdot g_{M_t},  \end{split} \\
\begin{split} &g_{M_0}(t) = g_{M_t}\cdot (M_{t-1}^P)^T,  \end{split} \\
\begin{split} &g_{M_0} = \partial{C_f}/\partial{M_0} = g_{M_0^P} + \sum_{i=1}^N{g_{M_0}(i)}, \end{split} \label{eq:flow_grad_1} \\
\begin{split} &f'_P(x) = 1\ if\ x > 0,\ else\ 0, \end{split} \\
\begin{split} &f'_I(x) = 1. \end{split}
\end{align}
The computing graph for $C_f$ composes of a main data path from $M_0$ through a series of MC iterations to $C_f$ and a set of side paths directly connecting $M_0$ to the input of expansion node. Therefore, the gradient of $C_f$ respecting $M_0$ is computed by summing all gradients respecting $M_0$ input to all expansion node, as illustrated in Equation \ref{eq:flow_grad_1}. In addition, to simplify the gradient computation, we set the threshold of pruning to be $0$, making it be equivalent to a ReLU operation. Thus, the inflation becomes identical mapping with a gradient of $1$. This trick will slightly increase the number of iterations for convergence but simplifies the gradient computation, leading to a faster training in general. In this testing phase, the threshold can be turned up for faster convergence. \par

\section{Detail Implementation of MCN for Scene Text Detection}

The architecture of MCN, inference flow and training flow are shown in Figure \ref{fig:sys_flow}. An MCN consists of a CNN backbone network inherited from a pretrained VGG-16 model. We remove all the fully-connected layers and output features of the \emph{conv5\_3} with $1/16$ resolution. For an input image size of $H\times W$, the \emph{conv5\_3} output is of size $H/16 \times W/16$. The \emph{conv5\_3} features are respectively fed to a Fore-/Background Subnetwork (FBN) and a Local Correlation Subnetwork (LCN). FBN detects multi-scale objects with a Feature Pyramid Network (FPN) \cite{lin2016feature} and a 2D-Recurrent Neural Network (2D-RNN). LCN predicts spatial and semantic correlation between adjacent image patches with stride of $16$. The objects' presence probability $P \in (0,1)^{H/16 \times W/16}$ and the local correlation measurements $S_1, S2$ and $S_3 \in (0,1)^{H/16 \times W/16}$ between current image patch and its three neighbors (bottom, right and left) produced by FBN and CSN respectively are translated into four flow maps $f_0, f_1, f_2$ and $f_3 \in \mathbb{R}^{+^{H/16 \times W/16}}$. A latticed Stochastic Flow Graph (SFG) is constructed from flow maps which is further described by a flow matrix $M_0$. By performing Markov Clustering on the SFG, we can group nodes that belongs to the same object together and generate instance-level bounding boxes based on Principle Component Analysis (PCA). \par

The MCN is end-to-end trainable with bounding box level labeling. As illustrated in Figure \ref{fig:sys_flow}, the ground-truth bounding boxes are converted to node-wise object mask $y_o \in \{0,1\}^{H/16 \times W/16\times 2}$ and flow label $y_f(i_m,j_m) \in \mathbb{R}^{\frac{H}{16}\times\frac{W}{16} \times (\frac{H}{16}\cdot \frac{W}{16})}$, which are used to compute the Object Loss $L_o \in \mathbb{R}^{H/16 \times W/16}$, Object Cost $C_o =\frac{1}{{H}/{16}\cdot {W}/{16}}\sum_{i_m}\sum_{j_m} L_o(i_m,j_m)$, Flow Loss $L_f \in \mathbb{R}^{H/16\cdot W/16}$ and Flow Cost $C_f \in \mathbb{R}$. The total cost $C_{total} \in \mathbb{R}$ is computed by summing $C_o$ and $C_f$ together.

\section{Experiments} \label{sess:experiment}

We evaluate the proposed model on three public scene text detection datasets, namely ICDAR 2013, ICDAR 2015 and MSRA-TD500, using the standard evaluation protocol proposed in \cite{wolf2006object,karatzas2015icdar,yao2012detecting}.
\subsection{Datasets}

\textbf{SynthText \cite{gupta2016synthetic}} contains over 800,000 synthetic scene text images. They are created by blending natural images with text rendered with random fonts, size, orientation, and color. It provides word level bounding box annotations. We only use this dataset to pretrain our model.

\textbf{ICDAR 2013 \cite{karatzas2013icdar}} is a dataset containing horizontal text lines. It has 229 text images for training and 223 images for testing.

\textbf{ICDAR 2015 \cite{karatzas2015icdar}} consists of 1000 training images and 500 testing images. This dataset features \emph{incidental} scene text images taken by Google Glasses without taking care of positioning, view point and image quality.

\textbf{MARA-TD500 \cite{yao2012detecting}} is a multilingual dataset focusing on oriented texts. It consists of 300 training images and 200 testing images.

\subsection{Experiment Details}

Our model is pre-trained on SynthText and finetuned on real datasets. It is optimized by the standard SGD algorithm with a momentum of $0.9$. Both training and testing images are resized to $512\times 512$. The batch size is set to $20$. In pretraining, the learning rate is set to $10^{-3}$ for the first 60k iterations, and decayed by a factor of $10$ for the rest 30k iterations. The finetuning on public benchmarks runs at learning rate of $10^{-5}$ with data augmentation proposed in \cite{liu2016ssd}. In testing, the threshold used for Pruning is set to $0.15$. Both the training and testing flows are implemented with TensorFlow \cite{abadi2016tensorflow} r1.1 on Dell Precision T7500 workstation with Intel Xeon 5600 processor, 40 GB memory and a NVIDIA GTX 1080 GPU.

\subsection{Detail Analysis}

\textbf{Baseline Comparison} We conduct an experiment to validate the performance gain is coming from the proposed framework. The baseline model (Local-link) predicts the fore/background and four local link scores between nodes to capture the local correlation information. The instance-level bounding boxes are generated by finding the maximum connected (by link scores) component on the foreground regions. Both the baseline model and the MCN model is constructed based on VGG-16 backbone, and we keep the number of parameter to be roughly equal. The performance is shown in Table \ref{tab:icdar13}. It concludes that our method is overall better than the baseline setting (Local-link). In local-link model, nodes between two text regions may be unexpectedly connected by undirected links, leading to a fusion of two individual text instances. Due to a directed flow prediction and a data-driven clustering mechanism, MCN greatly reduces unexpected connections and can provide more robust instance-level bounding box proposal. \par

\begin{table}[h]\renewcommand{\arraystretch}{1.0}
	\centering
	\caption{Analysis experiment on ICDAR 2013.}
	\label{tab:icdar13}
	\begin{tabular}{|l||l|l|l|}
		\hline
		& P ($\%$)    & R ($\%$)    & F ($\%$)    \\ \hline
		Local-link & 82.3 & 85.1 & 83.4 \\ 
		MCN  & \textbf{88.2} & \textbf{87.2} & \textbf{87.7} \\ \hline
	\end{tabular}
\end{table}

\begin{figure*}[]
	\centering
	\includegraphics[width=0.98\linewidth]{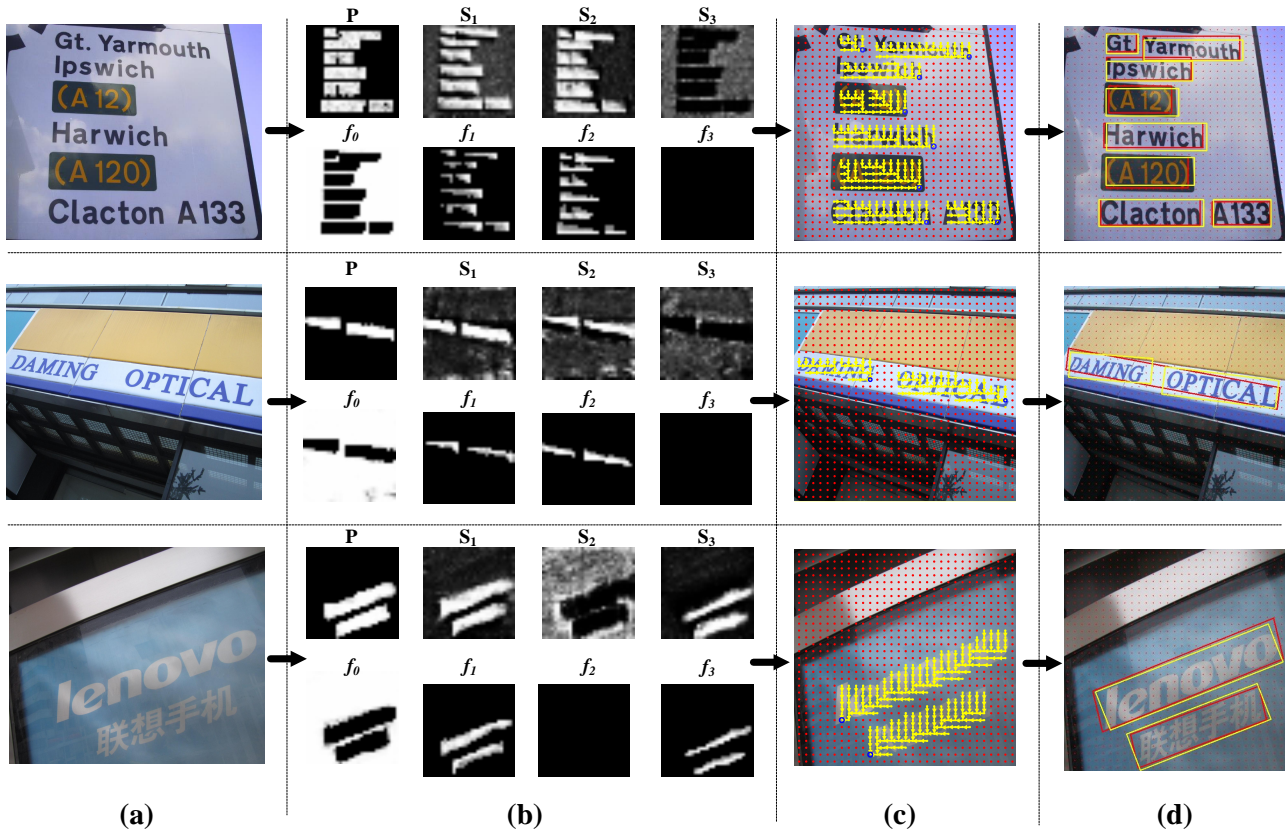}
	\caption{Flow profiling of detecting text with different orientations. (a) Input images; (b) Predicted objectness maps $P$, link maps $S_1, S_2, S_3$, and stochastic flows $f_0, f_1, f_2, f_3$; (c) Visualization of Markov Clustering with dominative flows marked in yellow arrows and attractors marked in blue points; (d) Generated bounding boxes (in yellow) based on predicted clusters and according ground-trues labeled in red boxes.}
	\label{fig:flow_demo}
\end{figure*}

\textbf{Profiling the MCN} Figure \ref{fig:flow_demo} visualizes the predicted flows by MCN. The input images with three orientations, horizontal, right-oblique and left-oblique are shown in Figure \ref{fig:flow_demo} (a). Figure \ref{fig:flow_demo} (b) profiles the activation maps including the object region prediction $P$, link scores $S_1$, $S_2$, $S_3$, and stochastic flows $f_0$, $f_1$, $f_2$, $f_3$. All the activation maps with size of $32\times 32$ originally are upsampled to $512\times 512$ for demonstration. According the activation map of $f_0$, $f_1$, $f_2$ and $f_3$, we draw the dominative flows and label the attractor on the input image, which is shown Figure \ref{fig:flow_demo} (c). In Figure \ref{fig:flow_demo} (d), the predicted bounding boxes and the ground-truth bounding boxes are labeled in yellow and red respectively. \par

\begin{table*}[t]\renewcommand{\arraystretch}{0.97}
\centering
\caption{Localization performance on ICDAR-13, ICDAR-15 and MSRA-TD500.}
\label{tab:full_cmp}
\scalebox{1}{
\begin{tabular*}{\textwidth}{@{\extracolsep{\fill}}|l||ccc||ccc||ccc|}
	\hline
	Dataset                                                 & \multicolumn{3}{c||}{ICDAR-13}                 & \multicolumn{3}{c||}{ICDAR-15}                 & \multicolumn{3}{c|}{MSRA-TD500}               \\ \hline
	Methods                           & \multicolumn{1}{c}{P}             & \multicolumn{1}{c}{R}             & \multicolumn{1}{c||}{F}             & \multicolumn{1}{c}{P}             &\multicolumn{1}{c}{R}             & \multicolumn{1}{c||}{F}             & \multicolumn{1}{c}{P}             & \multicolumn{1}{c}{R}             & \multicolumn{1}{c|}{F}             \\ \hline
	TextFlow \cite{tian2015text}                          & 0.85          & 0.76          & 0.80          & -             & -             & -             & -             & -             & -             \\ 
	Jaderberg \emph{et al.} \cite{jaderberg2016reading} & 0.89          & 0.68          & 0.77          & -             & -             & -             & -             & -             & -             \\ 
	Zhang \emph{et al.} \cite{zhang2016multi}             & 0.88          & 0.78          & 0.83          & 0.71          & 0.43          & 0.54          & 0.83          & 0.67          & 0.74          \\ 
	Gupta \emph{et al.} \cite{gupta2016synthetic}       & 0.92          & 0.75          & 0.83          & -             & -             & -             & -             & -             & -             \\ 
	Yao \emph{et al.} \cite{yao2016scene}               & -             & -             & -             & 0.72          & 0.59          & 0.65          & 0.77          & 0.75          & 0.76          \\ 
	TextBox \cite{liao2017textboxes}                      & 0.88          & 0.83          & 0.85          & -             & -             & -             & -             & -             & -             \\ 
	CTPN \cite{tian2016detecting}                         & \textbf{0.93} & 0.83          & \textbf{0.88} & 0.52          & 0.74          & 0.61          & -             & -             & -             \\ 
	SegLink \cite{shi2017detecting}                       & 0.88          & 0.83          & 0.85          & \textbf{0.72} & \textbf{0.77} & \textbf{0.75} & \textbf{0.86}          & 0.70          & 0.77          \\ 
	DeepReg \cite{he2017deep}      & 0.92              &0.81           &0.86        &\textbf{0.82}    &\textbf{0.80}     &\textbf{0.81}      & 0.77     &0.70    &0.74 \\ 
	\textbf{MCN}                                          & 0.88          & \textbf{0.87} & \textbf{0.88}          & 0.72 & \textbf{0.80}          & 0.76          & \textbf{0.88} &  \textbf{0.79} & \textbf{0.83} \\ \hline
\end{tabular*}
}
\end{table*}

On one hand, MCN shows the high accuracy in detecting text objects close to each other. As shown in objectness map $P$ at first row of Figure \ref{fig:flow_demo}, regions of multiple text objects merge together and we cannot generate bounding boxes directly from this map. The stochastic flows predicted by MCN captures the instance-level correlation and separate the merged regions into clusters. In some challenging casses with low quality $P$, the flow-based prediction can maintain good performance since an text object is jointly predicted by multiple nodes and their connections. On the other hand, the MCN method is flexible to handle text objects with different lengths and orientations. The orientation of an object is also represented by all flows within the object region jointly, resulting in a more accurate bounding box generation.   

\subsection{Performance Comparison}

\begin{figure*}[]
	\centering
	\includegraphics[width=0.93\linewidth]{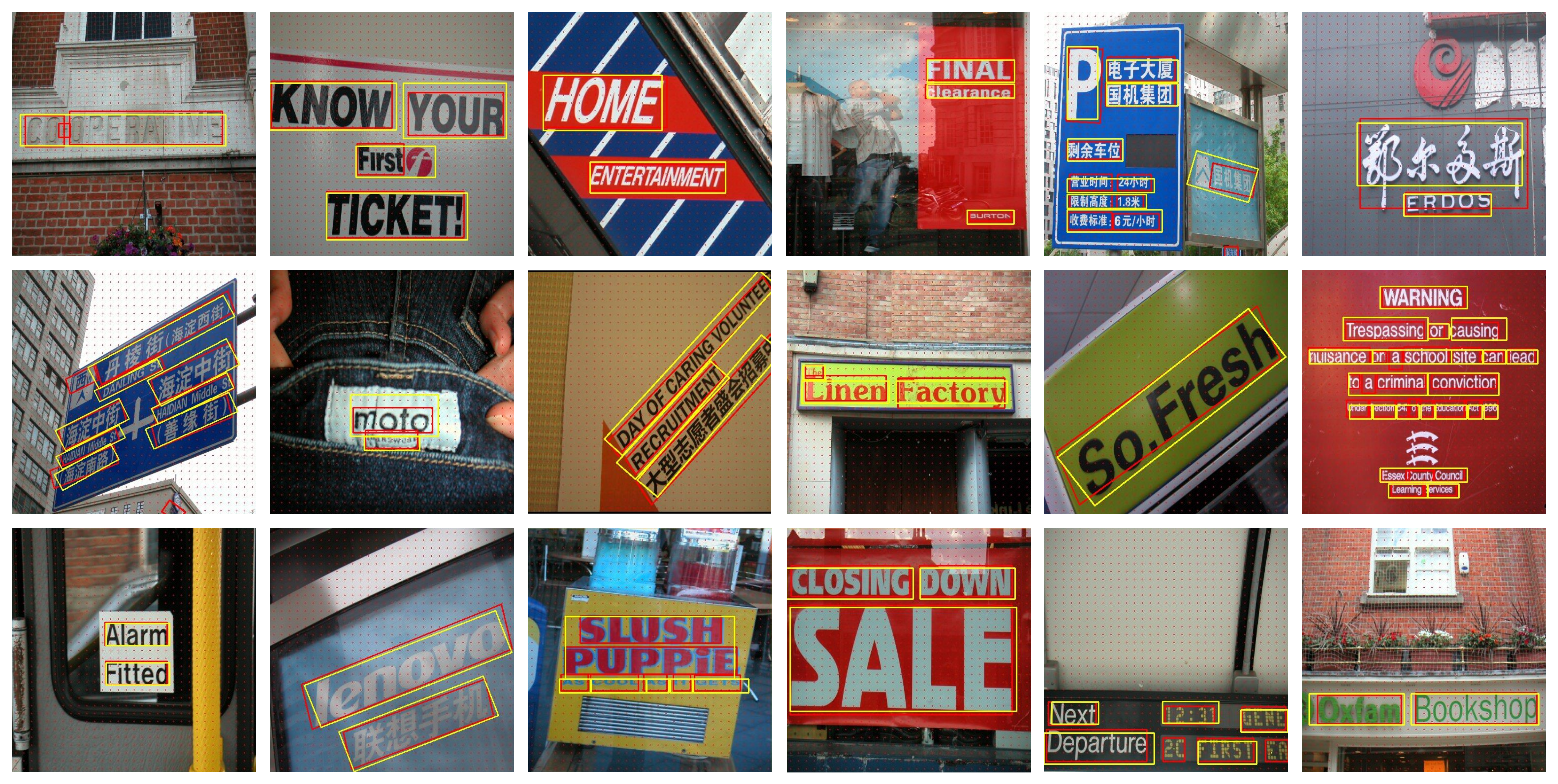}
	\caption{Example results on ICDAR 2013 and MSRA-TD 500 datasets. Text instances with multiple scales and orientations are detected. The predicted bounding boxes by our method are labeled in yellow and the ground-truth bounding boxes are labeled in red.}
	\label{fig:loc_results}
\end{figure*}

Table \ref{tab:full_cmp} compares our method with the published works on public datasets of scene text detection. On the ICDAR-13 dataset, our method reaches the state-of-art performance with precision of 0.88, recall of 0.87 and F-score of 0.88. On ICDAR-15, a slight performance drop is observed as compared to the existing text detection methods. Since most of the text objects are of size smaller than the node density ($16 \times 16$ pixel), the flows predicted for these objects are weak, leading to inaccurate object detection. But MCN achieves a new state-of-art performance on the MSRA-TD500 dataset. As shown in Table \ref{tab:full_cmp}, MCN outperforms the existing methods by a grate margin with precision of 0.88, recall of 0.79 and F-score of 0.83. Different from the ICDAR-13 that consists of only horizontal text objects, MSRA-TD500 contains large number of oblique and long text samples. The performance improvement in MSRA-TD500 shows that MCN is better at detecting multioriented text objects. \par

Figure \ref{fig:loc_results} demonstrates the bounding box prediction of MCN. The samples include both English and Chinese with different scales and orientations. The predicted bounding boxes are labeled in yellow and the ground-truths are labeled in red. As shown in Figure \ref{fig:loc_results}, MCN detects multilingual text objects with various scales and orientations robustly. The flow clustering framework supports the different bounding box geometry flexibly. Compared with the region proposal based scene text detection algorithms \cite{tian2016detecting,shi2017detecting,he2017deep,liao2017textboxes}, our method predicts more elaborated instance-level bounding boxes. As for the segmentation based methods \cite{yao2016scene,li2017unified}, our method involves much less heuristic operations. Both the flexibility and data-driven characteristics make MCN be superior to existing scene text detection methods. \par

\subsection{Speed}
In this section, we analyze the computation time of Markov Clustering. The Markov Clustering algorithm is implemented based on CUDA 8.0 with cuDNN 5 library \cite{nickolls2008scalable}. \par

We profile the computation time on Table \ref{tab:n_vs_msec}, as well as according precision, recall and F-score of bounding prediction with different $N$. In general, the computation time of Markov Clustering increases linearly with the increase of $N$. The detection performance also increases as $N$ increases, since the flow matrix $M_t$ requires sufficient number of iterations for convergence. Fortunately, it only takes few iterations for convergence to reach the best detection performance. As shown in Table \ref{tab:n_vs_msec}, MCN reaches its best performance with $N = 5$ and it takes only 0.86 ms to compute the clusters. This computing time is negligible when compared to the whole inference time of over 25 ms. \par

We also compare the inference speed in FPS with the recently proposed scene text detection methods on ICDAR-13 dataset. As shown in Table \ref{tab:fps}, our method achieves state-of-art performance and outperforms the existing methods with $1.5\times$ speedup. This is owing to flow-based method, which can tolerate inaccurate fore-/background prediction and thus maintains the same performance with less network parameters.

\begin{table}[] \renewcommand{\arraystretch}{0.9}
	\centering
	\caption{Number of iterations, detection performance, and runtime.}
	\label{tab:n_vs_msec}
	\scalebox{1}{
	\begin{tabular}{|c||c||ccc||c|}
		\hline
		\multicolumn{2}{|c||}{}                & \multicolumn{1}{|c}{P} & \multicolumn{1}{c}{R} & F    & time (ms) \\ \hline
		& N=1   & 0.7                    & 0.34                   & 0.46 & 0.32      \\ \cline{2-5} \cline{6-6} 
		& N=2   & 0.72                   & 0.65                   & 0.68 & 0.51      \\ \cline{2-5} \cline{6-6} 
		& N=3   & 0.78                   & 0.71                   & 0.74 & 0.67      \\ \cline{2-5} \cline{6-6} 
		MCN                           & \textbf{N=4}   & \textbf{0.85}                   & \textbf{0.85}                   & \textbf{0.85} & \textbf{0.86}      \\ \cline{2-5} \cline{6-6} 
		& \textbf{N=5}   & \textbf{0.88}                   & \textbf{0.87}                   & \textbf{0.88} & \textbf{1.05}      \\ \cline{2-5} \cline{6-6} 
		& N=6   & 0.88                   & 0.87                   & 0.88 & 1.23      \\ \cline{2-5} \cline{6-6} 
		& N=7   & 0.88                   & 0.87                   & 0.88 & 1.41      \\ \cline{2-5} \cline{6-6} 
		& N=8   & 0.88                   & 0.87                   & 0.88 & 1.60      \\ \cline{1-5} \cline{6-6} 
	\end{tabular}
	}
\end{table}
\begin{table}[] \renewcommand{\arraystretch}{1}
	
	\centering
	\caption{FPS comparison on ICDAR-13 with input size of $512\times 512$.}
	\label{tab:fps}

	\begin{tabular}{|l||ccc||c|}
		\hline
		Methods                                                 & P                         & R                                  & F                         & FPS                                \\ \hline
		Yao \cite{yao2012detecting} & 0.63           &0.63       &0.60 &0.14 \\ \hline
		Gupta \cite{gupta2016synthetic} &92.0 &75.5 &83.0 &15 \\ \hline
		TextBox \cite{liao2017textboxes}                      & 0.88                      & 0.83                               & 0.85                      & 20.6                               \\ \hline
		CTPN \cite{tian2016detecting}                         & \textbf{0.93}             & 0.83                               & \textbf{0.88}             & 14.2                               \\ \hline
		SegLink \cite{shi2017detecting}                       & 0.88                      & 0.83                               & 0.85                      & 20.6                               \\ \hline
		\textbf{MCN}                                  & 0.88                      & \textbf{0.87}                      & \textbf{0.88}                      & 34                               \\ \hline
	\end{tabular}

\end{table}

%

\section{Conclusion}

We present a novel Markov Clustering Network (MCN) for scene text detection. We treat the object detection problem as a graph-based clustering problem and develop a end-to-end trainable model for flexible scene text detection. MCN shows superiority in the sense of accuracy, robustness and speed. MCN outperforms the existing scene text detection algorithms in detecting multiscale and multioriented text objects. It also achieves $1.5\times$ speedup in comparison with the state-of-art algorithm. Our method is complementary to the existing top-down methods. Applying the extra top-down information to further improve the detection performance will be  consider as future research extension.

\newpage
\clearpage

{\small
	\bibliographystyle{ieee}
	\bibliography{egbib}
}

\newpage
\clearpage

\appendix
\section{Appendices}
\balance
\subsection{Bounding Box Generation}
Given a vertex $V_m$ in a cluster $\Psi$, we compute their coordinates in the input image $\omega_m = (i_m ,j_m)\cdot\mu_{stride} + \mu_{offset}$, where $\mu_{stride} = 16$ and $\mu_{offset} = 8$. Then the bounding box size and orientation of each cluster are computed based on Principle Component Analysis (PCA). Given a set of coordinates $\omega = \{\omega_m|m=1,2,\cdots\}$ of a cluster , we compute the its eigenvectors $\theta_1$ and $\theta_2$ as well as the corresponding eigenvalues $\lambda_1$ and $\lambda_2$. The coordinates of the four corners of the bounding box is computed by:

\begin{equation}
\begin{split}
c_1 & = A(\lambda_1\cdot\theta_1 + \lambda_2\cdot\theta_2) + \phi \\
c_2 & = A(\lambda_1\cdot\theta_1 - \lambda_2\cdot\theta_2) + \phi \\
c_3 & = A(-\lambda_1\cdot\theta_1 - \lambda_2\cdot\theta_2) + \phi \\
c_4 & = A(-\lambda_1\cdot\theta_1 + \lambda_2\cdot\theta_2) + \phi 
\end{split}
\end{equation}
where $\phi$ is the center of the cluster and $A$ denotes the scaling factor which is set to $1.75$. 

\subsection{From Image to Stochastic Flow}

Crucially, accurate object detection relies on correct flow prediction. In MCN, the flows $f_0$, $f_1$, $f_2$ and $f_3$ are the outputs of the Flow Mapping Layer (FML) with regional object probability $P$ and correlation measurement $S_1$, $S_2$ and $S_3$ as inputs. $P$ is generated by the Fore-/Background Network (FBN), while $S_1$, $S_2$ and $S_3$ are output by Local Correlation Network (LCN). Both FBN and LCN are starting at the \emph{conv5\_3} of VGG-16 pretrained network.\par

\subsubsection{Fore-/Background Network}
\begin{figure*}[]
	\centering
	\includegraphics[width=0.9\linewidth]{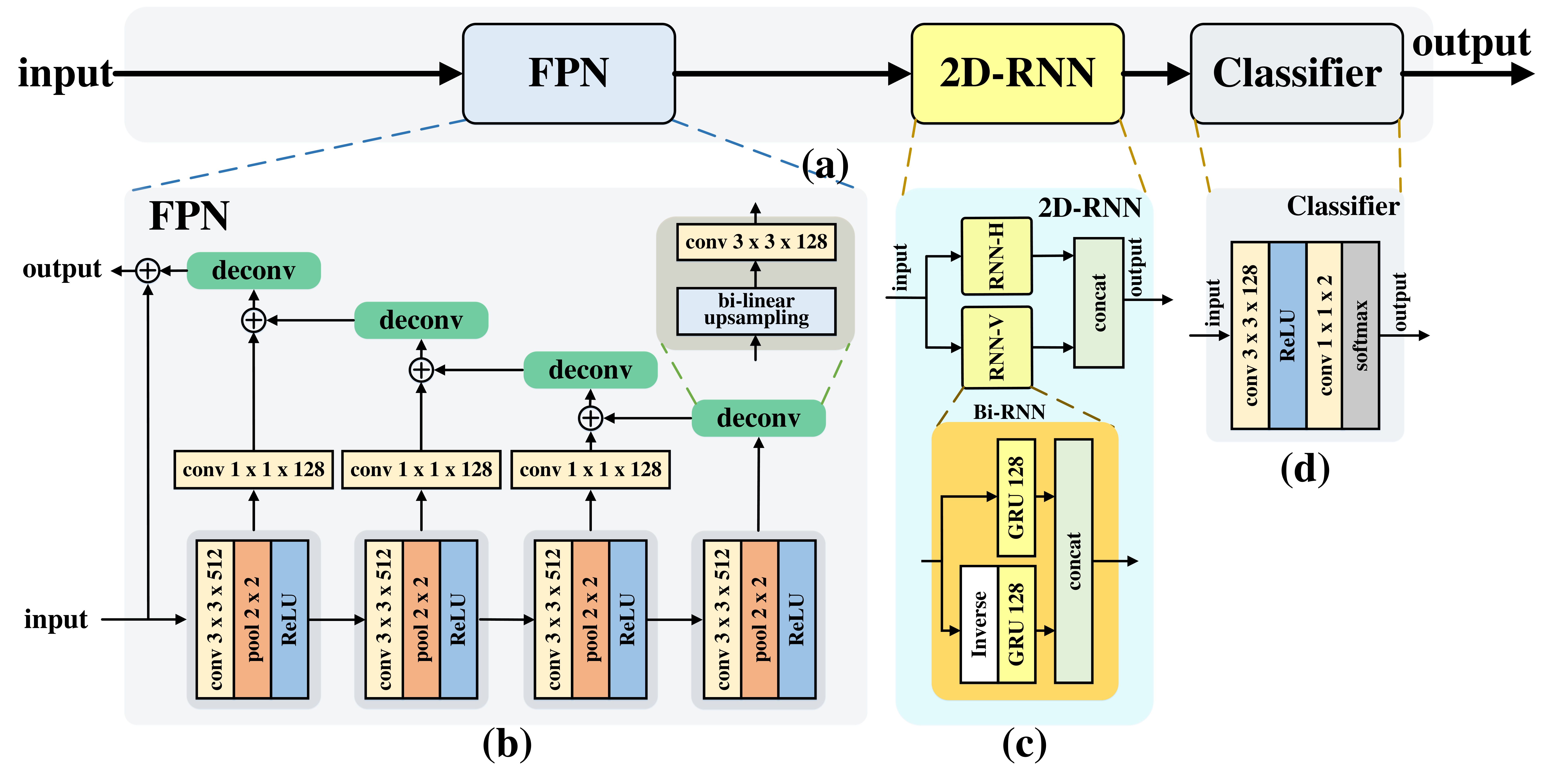}
	\caption{(a) Architecture of Fore-/Background Network (FBN); (b) Feature Pyramid Network (FPN) fusing feature maps with different resolutions; (c) 2-dimensional Recurrent Neural Network (2D-RNN) encoding contextual representations; (d) Regional objectness classifier predicting presence of an object with stride of $16\times 16$ pixels.}
	\label{fig:objectnet}
\end{figure*}

As shown in Figure \ref{fig:objectnet} (a), the Fore-/Background Network is an FPN-based network \cite{lin2016feature} with spatial recurrent components and \emph{softmax} output to predict the object score $P \in (0,1)^{H_{1/16}\times W_{1/16}}$. The output of \emph{conv5\_3} is further processed by a Feature Pyramid Network (FPN) and a 2-dimensional Recurrent Neural Network (2D-RNN) successively. In FPN shown in Figure \ref{fig:objectnet} (b), input with size of $H/16 \times W/16$ is processed by four convolutional blocks with $2\times 2$ pooling layers to obtain additional feature maps with resolution of $1/32$, $1/64$, $1/128$ and $1/256$. These feature maps together with the input are fused to resolution of $1/16$ by deconvolution consisted of layer-wise addition, bilinear upsampling and convolution. By fusing features with different resolution in a pyramid manner, our method have larger capacity to detect multiscale objects with less parameters. Subsequently, the output of FPN is fed to an 2D Recurrent Neural Network (2D-RNN) before region-based classification. We consider a spatial feature map as a 2D sequence which can be directly analyzed by a 2D-RNN. The structure of the proposed 2D-RNN is shown in Figure \ref{fig:objectnet} (c). A 2D-RNN is composed of two Bidirectional RNNs (RNN-H and RNN-V), which are applied to the rows and columns of the input feature map independently. As shown in Figure \ref{fig:objectnet}, the outputs of 2D-RNN is constructed by concatenating two feature maps produced by RNN-H and RNN-V with size of $H_{1/16}\times W_{1/16}$ along depth axis. Finally, a region-based classification is performed on the output feature map by a 2-layer convolutional network with \emph{softmax} output, Figure \ref{fig:correlatnet} (d).

\subsubsection{Local Correlation Subnetwork}

\begin{figure}[]
	\centering
	\includegraphics[width=0.9\linewidth]{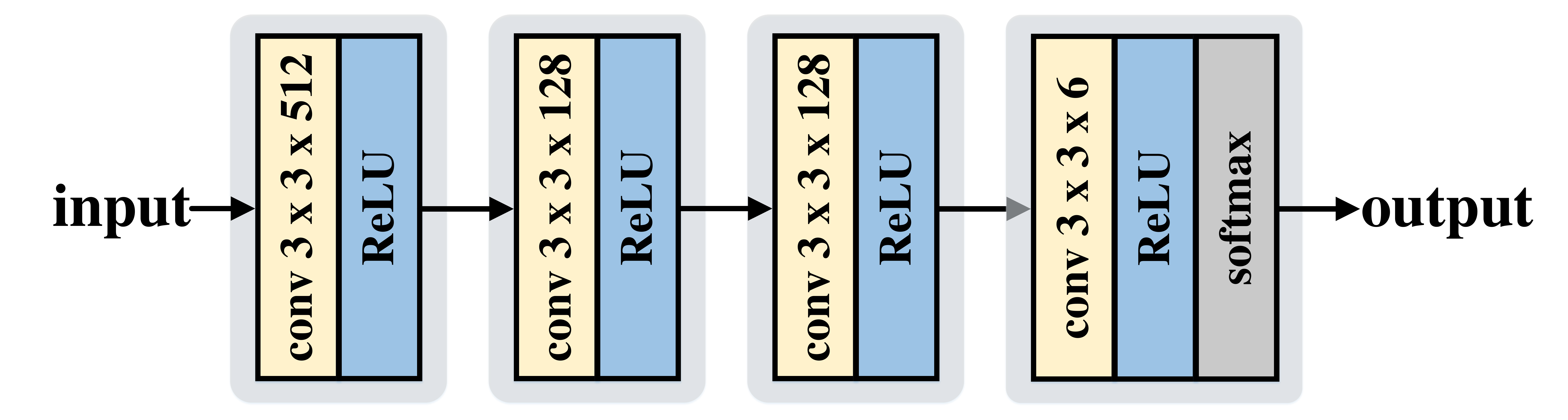}
	\caption{Local Correlation Network (LCN) with four convolution blocks and a softmax layer outputs correlation measurements $S_1$, $S_2$ and $S_3$ between current anchor and its three neighbors (bottom, right and left).}
	\label{fig:correlatnet}
\end{figure}

To predict the semantic and spatial correlation between adjacent subregions, we build another subnetwork with additional four convolutional blocks and a \emph{softmax} classifier starting at \emph{conv5\_3}, shown in Figure \ref{fig:correlatnet}. The network outputs three correlation measurements $S_1$, $S_2$ and $S_3$ $\in (0,1)^{H_{1/16}\times W_{1/16}}$ representing the semantic and spatial correlation between current anchor and its three neighbors (bottom, right and left) respectively. As the \emph{conv5\_3} features is corresponding to subregions of input image with stride of $16$, the LCN is actually measuring the correlation among these overlapping subregions. Together with output of objetness network $P$, $S_1$, $S_2$ and $S_3$ are mapped to the Stochastic Flow $f_0$, $f_1$, $f_2$ and $f_3$ by Flow Mapping Layer (FML).

\subsubsection{Flow Mapping Layer}
The Flow Mapping Layer (FML) is point-wise non-linear function with input of $P$, $S_1$, $S_2$ and $S_3$ and output of $f_0$, $f_1$, $f_2$ and $f_3$. The mapping is shown below: 

\begin{equation}
\label{eq:f0}
f_0 = e^{-\alpha[1-\mu(1-P)]\cdot[S_1^2+S_2^2+S_3^2]}
\end{equation}
\begin{equation}
\label{eq_f1}
f_1 = (1-f_0)\cdot \frac{S_1}{S_1 + S_2 + S_3}
\end{equation}
\begin{equation}
\label{eq_f2}
f_2 = (1-f_0)\cdot \frac{S_2}{S_1 + S_2 + S_3}
\end{equation}
\begin{equation}
\label{eq_f3}
f_3 = (1-f_0)\cdot \frac{S_3}{S_1 + S_2 + S_3}
\end{equation}
\begin{equation}
\label{eq:mu}
\mu(x) = \frac{1}{1+e^{-\beta(x-\gamma)}}.
\end{equation}
Here, $f_0$ is actually the transition probability of \emph{self-loop}, which is controlled by the likehood of background ($1-P$) and the correlation measurement between current vertex and its neighbors ($S_1$, $S_2$ and $S_3$). It is designed to be weak for vertices within the same object region and to be strong for a vertex which corresponds to the background or is just the attractor of a cluster. This behavior is realized by firstly measuring the correlation intensity ($S_1^2+S_2^2+S_3^2$) modulated by an \emph{on-off} function $\mu(x)$, and then projecting it to the exponential space. $\mu(x)$ is parameterized by trainable variables $\alpha$, $\beta$ and $\gamma$. It takes $1-P$ as input and produces an on-off signal to control $f_0$. It will disables the effect of $S_1$, $S_2$ and $S_3$ and drive $f_0$ approaching to 1 when a vertex is in the background region. Accordingly, the values of $f_1$, $f_2$ and $f_3$ will be small, making all the background vertices to be isolated. In the object region, the correlation intensity $S_1$, $S_2$ and $S_3$ take control of $f_0$ since $1-P$ is small. In this case, $f_0$ will be large if weak correlation is measured and the vertex will become the attractor of a cluster. Otherwise, the vectices belongs to the same object region will be connected through $f_1$, $f_2$ and $f_3$ and the flows of a cluster will end at the attractor.\par
\end{document}